\documentclass[letterpaper]{article} 
\usepackage{aaai2027} 
\usepackage[hyphens]{url}  
\usepackage{graphicx} 
\usepackage{natbib}  
\usepackage{caption} 
\usepackage{algorithm}
\usepackage{algorithmic}

\usepackage{newfloat}
\usepackage{listings}
\DeclareCaptionStyle{ruled}{labelfont=normalfont,labelsep=colon,strut=off} 
\floatstyle{ruled}
\newfloat{listing}{tb}{lst}{}
\floatname{listing}{Listing}

\usepackage{booktabs}

\usepackage{amsmath}
\usepackage{multirow}
\usepackage{makecell}
\usepackage{amssymb}
\usepackage{pifont}
\usepackage{cleveref}
\usepackage{multicol}

\newcommand{\up}[1]{{\scriptsize{(↑#1)}}}
\newcommand{\down}[1]{{\scriptsize{(↓#1)}}}
\newcommand{\cmark}{\ding{51}} %
\newcommand{\xmark}{\ding{55}} %

\title{ReMem: Streaming Video Understanding With Long Context Retention}
\author {
    Li Yiheng\textsuperscript{\rm 1}\equalcontrib,
    He Xu\textsuperscript{\rm 1}\equalcontrib,
    Wang Shaobo\textsuperscript{\rm 2},
    Shao Ling\textsuperscript{\rm 3},
    Lu Shijian\textsuperscript{\rm 1}\corresponding
}
\affiliations {
    \textsuperscript{\rm 1}Nanyang Technological University\\
    \textsuperscript{\rm 2}Shanghai Jiaotong University\\
    \textsuperscript{\rm 3}University of the Chinese Academy of Sciences\\
}

\begin{document}

\maketitle
\thispagestyle{plain}
\pagestyle{plain}
\begin{abstract}
Despite their impressive performance on a wide range of video understanding tasks, current Vision Language Models (VLMs) are predominantly designed for offline scenarios and struggle to handle online streaming videos that demand low latency response. Several studies have explored memory and token compression strategies in an attempt to adapt offline VLMs for streaming video understanding tasks. However, through our probing experiment, we identify that most existing works tend to progressively lose long context information as length of input stream increases. To address this, we propose ReMem, a novel training-free adaptation technique that enables VLMs to process streaming videos of arbitrary lengths while improving their long context information retention capability. ReMem exploits memory from two perspectives, implemented as two core components. The Streaming Context Memory (SCM) continuously compresses historical context with query-independent attention. The Retrieved Vision Memory (RVM) then retrieves the most salient, query-relevant context from memory to augment the VLM's input. Comprehensive experiments demonstrate that the proposed ReMem achieves state-of-the-art (SOTA) performance across a variety of widely used benchmarks, spanning both streaming video  and general long video understanding tasks.
\end{abstract}
\section{Introduction}
\label{sec:intro}
Vision Language Models (VLMs) have demonstrated remarkable advancements in a wide range of multi-modal tasks, from single image question-answering \cite{fu2025mme, chen2024we} to complex multi-turn dialogue \cite{liu2024mmdu, li2023scigraphqa} and long video understanding \cite{fu2025videommefirstevercomprehensiveevaluation, wu2024longvideobenchbenchmarklongcontextinterleaved}. However, this progress has largely been confined to offline settings, where models presume access to the entire visual input before generating a response \cite{bai2025qwen2,wang2024qwen2,Qwen-VL, chen2024internvl, chen2024far, chen2024expanding, zhang2024videoinstructiontuningsynthetic, apollo}. This assumption might be impractical for many real-time applications where visual data is provided as a continuous stream. Applications such as autonomous driving \cite{shao2024lmdrive} require fast response and expect low latency from VLMs. Developing models that can operate effectively in these online, streaming environments remains a challenge.

Recent efforts have sought to adapt VLMs to the streaming video setting by lowering the latency of the answer generation process \cite{zhang2025flashvstream, di2025rekv, wang2025streambridge}. One key strategy is token compression \cite{yang2025visionzip, tao2025dycoke, chen2024image}, which reduces the input sequence length to speed up VLM inference. 
In addition, memory-based modules \cite{song2024moviechat, song2025moviechat+, wang2024videollamb} are often used in conjunction with compression. These modules store long context information from older segments of the input video stream, allowing relevant context to be retrieved later when a user query arises.

While such methods effectively reduce the token length for VLM transformer operations \cite{vaswani2017attention} and speed up the inference process, they often come at the cost of long context retention. We demonstrate this tradeoff using a variant of visual Needle-In-A-Haystack (NIAH) probing experiment \cite{wu2024visual}. As shown in \cref{fig:vhs_comparison}, our results reveal that as the haystack size increases, existing streaming models such as FluxMem \cite{xie2026fluxmemadaptivehierarchicalmemory} and OASIS~\cite{liang2026oasis} suffer a more pronounced drop in retrieval accuracy than the Qwen2.5VL~\cite{bai2025qwen2} base model. This indicates that these methods increasingly fail to preserve crucial contextual information, progressively losing key details as visual sequences grow longer. In contrast, our proposed method, ReMem, consistently surpasses these evaluated models by effectively capturing and retaining key information despite the increasing context size. 

This limitation motivated us to design a streaming VLM adaptation that retains long-term context while maintaining a short input token length for efficiency. We introduce ReMem, a training-free VLM adaptation strategy that seamlessly enables offline pre-trained VLMs to process online streaming data while preserving long-term context. ReMem consists of two main components: Streaming Context Memory (SCM) and Retrieved Vision Memory (RVM). SCM maintains a fixed-length memory module that continuously condenses historical visual information using attention-based compression. RVM then augments the model's input by retrieving the most relevant, query-specific vision tokens. Our experimental results show that ReMem achieves significant performance gains on streaming video benchmarks, especially on tasks requiring contextual understanding.
\newpage
In summary, our main contributions are:

1. We use visual NIAH probing experiments to demonstrate that existing memory based streaming VLM designs suffer from the progressive loss of long-term context. 

2. We introduce ReMem, a novel, training-free adaptation strategy that equips existing offline VLMs to handle streaming video of arbitrary length while simultaneously improving their ability to retain long-context information. 

3. We conduct comprehensive experiments on streaming and general long video understanding benchmarks. The results demonstrate that ReMem significantly outperforms baselines and competing streaming methods, particularly on streaming tasks, thereby validating the effectiveness of our design.
\section{Related Works}

\subsection{Vision Language Models}
VLMs have developed tremendously since their early forms \cite{liu2023llava, alayrac2022flamingovisuallanguagemodel,dai2023instructblipgeneralpurposevisionlanguagemodels,zhu2023minigpt4enhancingvisionlanguageunderstanding}. However, the basic structure of VLMs largely remains the same. They typically consist of a vision encoder \cite{dosovitskiy2021imageworth16x16words}, a connector \cite{liu2023llava}, and an LLM backbone \cite{peng2023instruction}. They function by concatenating visual features with text embeddings to auto-regressively generate outputs \cite{zhu2025internvl3exploringadvancedtraining, yu2025minicpmv45cookingefficient}. This foundational design is almost exclusively intended for offline settings, where the entire visual input is provided at once. Consequently, the architecture is inherently constrained by the LLM's fixed context window, making these VLMs ill-suited for real-world streaming applications where visual data arrives continuously and must be processed with low latency.

\subsection{Streaming Video Understanding}
Streaming video understanding aims to transition traditional offline VLMs toward practical online scenarios, where models must process video streams continuously rather than relying on entire preloaded video sequences. In such settings, VLMs are required to maintain long context consistency, reason over evolving visual evidence, and generate responses interactively and proactively in real time.
Early works \cite{degeest2016onlineactiondetection,xu2019temporalrecurrentnetworksonline,xu2021longshorttermtransformeronline,zhao2022realtimeonlinevideodetection,girdhar2021anticipativevideotransformer} in the action detection and anticipation domains   address this constraint by causally processing the historical video inputs together with the current observation.
More recent approaches tailor Video-LLMs for online use by modifying training objectives, memory mechanisms, or response strategies. For instance, VideoLLM-Online~\cite{VideoLLM-online} introduces an online training objective and stream-aware decoding, but heavily compresses each frame into only a few visual tokens, limiting fine-grained perception.
FlashVStream~\cite{zhang2025flashvstream} employs temporal–spatial memory and sequential retrieval for real-time reasoning over long streams, whereas Dispider~\cite{qian2025dispiderenablingvideollms} disentangles perception, decision, and reaction to support active interaction.
LiveVLM~\cite{ning2025livevlmefficientonlinevideo}, a training-free framework designed specifically for streaming, online video understanding, constructs a streaming-oriented KV cache that retains useful details while pruning redundant history.
TimeChat-Online~\cite{yao2025timechatonline80visualtokens} tackles visual redundancy in streaming inputs via a Differential Token Drop (DTD) module that preserves salient temporal changes while discarding static frames.
StreamBridge-7B~\cite{wang2025streambridge}, a simple yet effective framework that turns offline Video-LLMs into streaming-capable models, relies on a memory buffer with round-decayed compression and a lightweight decoupled activation module to enable multi-turn, proactive responses.
Complementary efforts such as Lion-FS~\cite{li2025lionfsfastslow} and LiveCC~\cite{chen2025livecclearningvideollm} explore scalable data/optimization strategies and speech-conditioned streaming, while MMDuet~\cite{wang2024videollmknowsspeakenhancing} and ViSpeak~\cite{fu2025vispeakvisualinstructionfeedback} introduce task-specific heads or duet-style interaction formats to decide when the model should speak.
Despite these advances, most methods either require training, modify the model architecture, or still struggle to maintain contextual fidelity over long visual sequences so that our training-free ReMem is proposed.

\subsection{Visual Token Compression}
Query-based compression strategies, like the Q-former \cite{li2023blip2bootstrappinglanguageimagepretraining}, select features most salient to the current user prompt. This approach is unsuitable for streaming video, which often involves multi-turn dialogue. Compressing visual tokens based on a single, specific query would irreversibly discard information that, while irrelevant to the current prompt, may be essential for answering subsequent user queries.

Due to the information sparsity in video tokens \cite{zhang2024sparsevlm, yao2025timechat}, some compression strategies focus on reducing the spatial and temporal redundancies of visual tokens \cite{zhang2025flashvstream, yao2025timechat}. However, the resulting compressed visual tokens differ from the model's original training data distribution and deviate from its learned inductive bias. Consequently, additional model fine-tuning is required to adapt to this new input token distribution.

To overcome this fine-tuning requirement, attention score based compression strategies \cite{yang2025visionzip, chen2024image, tao2025dycoke, wang2025adaretakeadaptiveredundancyreduction,yang2025streammemqueryagnostickvcache} select visual tokens or kV-cache vectors based on the model's own attention patterns. By retaining only the tokens with high attention scores, this approach creates a compressed token set that is less prone to out-of-distribution behavior and can maintain model performance without additional training. Given the effectiveness and resource-efficiency of these attention-based compression methods, we follow a similar design principle for our vision token compression module.

\subsection{Memory Based VLM}
To overcome the limitations of compression strategies and the restriction to fixed-length video inputs, researchers have proposed memory based approaches. Memory mechanisms are widely applied across long sequence modelling tasks, including forecasting, recommendation systems, and video object tracking. In the context of video understanding, various designs have been explored: ReKV\cite{di2025streamingvideoquestionansweringincontext}introduces a compressed KV-cache memory to retain earlier video information; MovieChat\cite{song2025moviechat+} adopts a dual memory framework with long-term and short-term memory;  However, a common drawback of these methods is the lack of an upper bound on memory size in at least one of their modules. As a result, when processing unbounded streaming videos, the accumulated token length eventually exceeds the hardware’s maximum capacity.

Other memory based solutions place explicit limits on memory usage. Flash-VStream\cite{zhang2025flashvstream} leverages separate memory modules for temporal and spatial modeling. InfiniPlotV \cite{kim2025infinipot} employs a memory compression strategy that enforces a fixed token length regardless of input size, while StreamBridge \cite{wang2025streambridge} introduces a round-decay mechanism that progressively merges earlier memory contexts with adjacent frames to maintain a constant memory footprint. However, such strategies inevitably merge or compress important visual features, particularly from earlier frames, leading to loss of long context information. In contrast, we propose an additional retrieval module that selectively extracts salient information from the memory buffer without merging or compressing critical details. The retrieved features augment our fixed-size memory, thereby mitigating information loss caused by progressive memory updates.

\section{Probing Experiment}
\label{probing experiment1}
Memory based streaming VLMs, such as \cite{zhang2025flashvstream, wang2025streambridge}, are designed to process videos of arbitrary length by maintaining a fixed-size context memory, which is continuously updated as new frames arrive. While this design is memory-efficient and achieves low latency, we hypothesize that it comes at the cost of long context retention, where crucial information from earlier in the input stream is progressively diluted or overwritten by newer, incoming data. 

\begin{figure}
    \centering
    \includegraphics[width=1.0\linewidth]{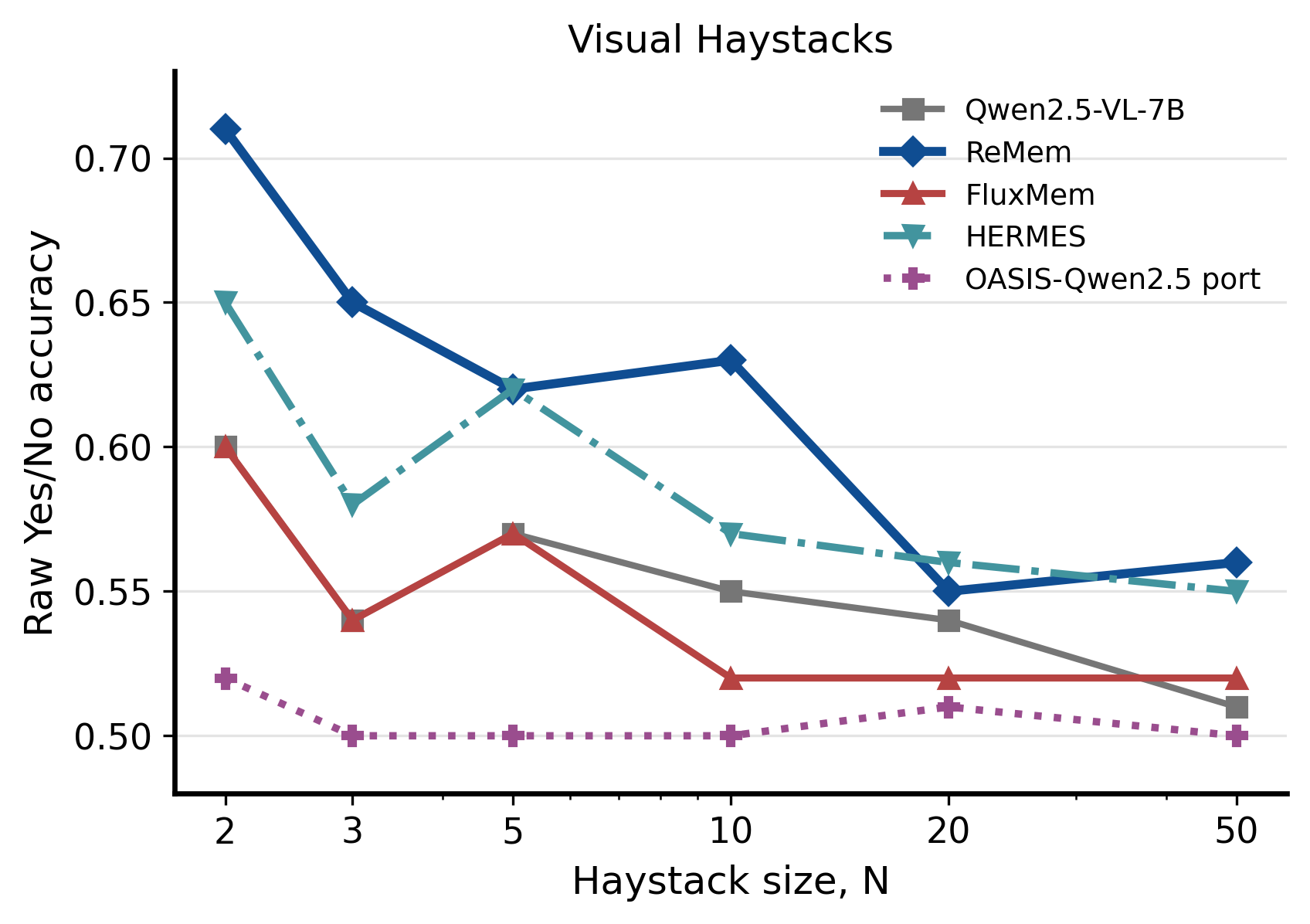}
    \caption{Comparison of visual haystacks retrieval performance among Qwen2.5VL base model~\cite{bai2025qwen2}, FluxMem~\cite{xie2026fluxmemadaptivehierarchicalmemory}, HERMES \cite{zhang2026hermeskvcachehierarchical}, OASIS \cite{liang2026oasis} and our method. As the haystack size increases, ReMem maintains more stable performance in long visual sequences, indicating that crucial contextual information is better preserved.} 
    \label{fig:vhs_comparison}
\end{figure}

To validate this hypothesis, we conduct a probing experiment using the Visual NIAH \cite{wu2024visual} dataset. The task requires the VLM to answer a question where the correct answer is based solely on a "needle" image placed at the very beginning of the input sequence. We then progressively increase the context length by appending irrelevant "haystack" images. We conduct this experiment on both variants of the FlashVstream model \cite{zhang2025flashvstream}, with the results presented in \cref{fig:vhs_comparison}.

As shown in \cref{fig:vhs_comparison}, both the Qwen2.5-VL-7B \cite{bai2025qwen2} base model and memory-augmented streaming VLMs suffer from accuracy degradation as the input context lengthens. Our results reveal that as the haystack size increases, existing streaming models such as FluxMem \cite{xie2026fluxmemadaptivehierarchicalmemory} and OASIS~\cite{liang2026oasis} suffer a more pronounced drop in retrieval accuracy. Their performance quickly drops to scores approaching a random guess (0.50) on this yes/no task.

This steep decline strongly suggests that the underlying context retention capability is compromised by the memory mechanisms employed in these models. As their memory modules continuously update to process longer input sequences, crucial visual information becomes progressively diluted by newer context, ultimately eroding their long-term memory. In contrast, our proposed method, ReMem, consistently surpasses these evaluated models by effectively capturing and retaining key information despite the increasing context size.

\section{Methodology}

Motivated by our probing experiments, we propose ReMem, a novel strategy that enables offline pretrained VLMs to process streaming video while preserving their long context understanding capabilities. ReMem consists of two core components:

\begin{itemize}
  \item Streaming Context Memory (SCM): This module condenses all incoming visual information into a fixed-length memory unit via attention-based visual token compression.
  \item Retrieved Vision Memory (RVM): This module enhances contextual features by extracting the most query-relevant visual tokens and re-injecting them into the model's input.
\end{itemize}

\begin{figure*}[t]
    \centering
    \includegraphics[width=\linewidth]{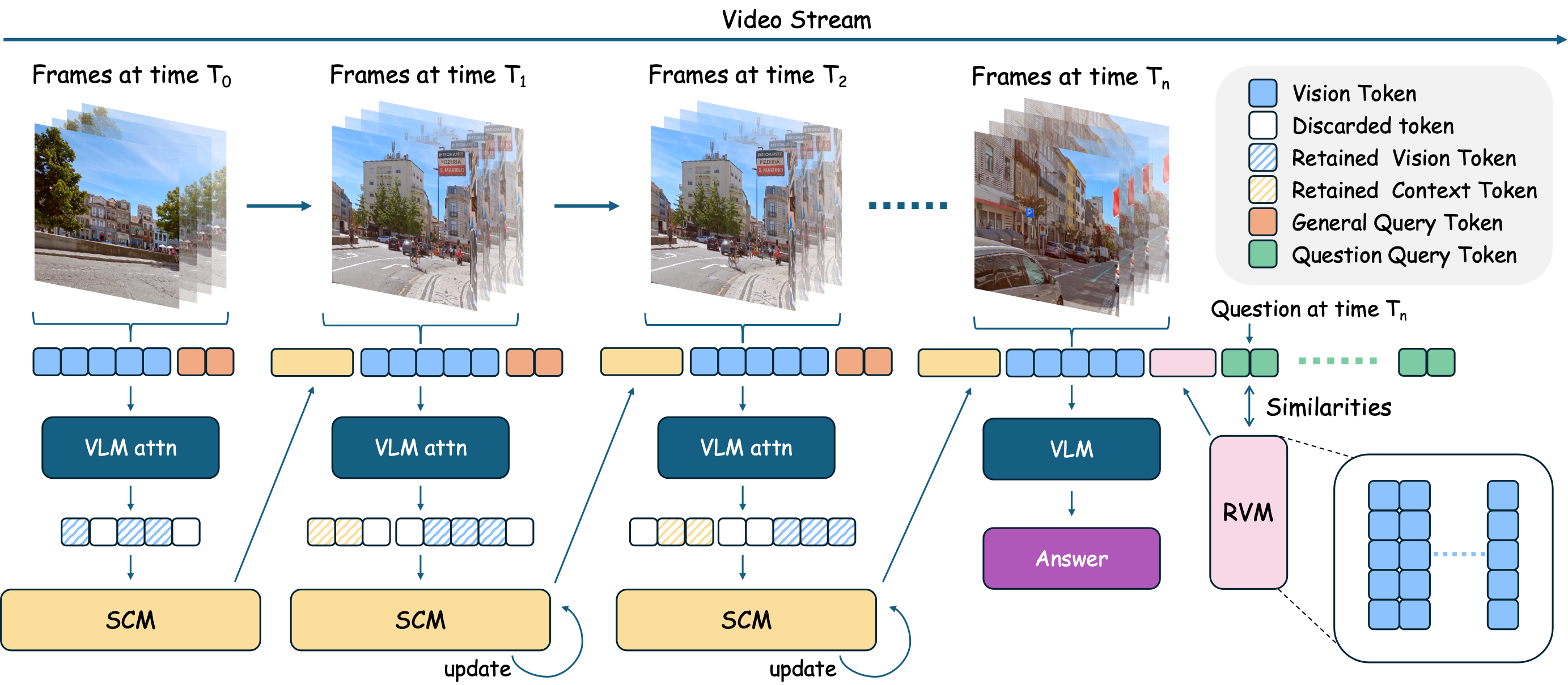}
    \caption{\textbf{Overview of the proposed ReMem framework.} The video stream is segmented to video chunks continuously.
    At each time step, the VLM attends to visual and contextual tokens, selectively storing informative representations into the SCM.
    Upon receiving a specific question query, we score each memory slot by cosine similarity, select the top vision tokens from the RVM, and integrate them with the ongoing context, yielding a compact representation. }
    \label{fig:main_method}
\end{figure*}

\subsection{Preliminary}

We define the streaming video task over a sequence of discrete time steps $T_i$, for $i \in \{0, 1, \ldots, n\}$. At each time step $T_i$, a new clip of video frames of length $C$ arrives from the stream. We can control the number of frames per video clip by setting different video sampling frequencies. These $C$ frames pass through the VLM's vision encoder and projection layer to produce a corresponding set of vision tokens. We denote this token segment, corresponding to the frames at $T_i$, as $V_i$.

\subsection{Streaming Context Memory}

As shown in \cref{fig:main_method}, SCM is a fixed-length memory storage that iteratively updates as new video segments enter the model. 

This update process is defined in \cref{eq:streaming_context}. We employ General Query Compression (GQC) as the compression strategy, which we elaborate in section below. For the initial time step $T_0$, the SCM tokens $S_0$ are initialised by compressing the first segment of vision tokens $V_0$. For all subsequent time steps $T_i$, the new memory state $S_i$ is derived by combining and compressing the previous context tokens $S_{i-1}$ and the current vision tokens $V_i$, with compression ration of $r$. The maximum size of SCM can be controlled by fixing $r$.

This recursive design ensures that $S_i$ contains condensed information from all vision tokens $V_0$ up to $V_i$ within a fixed-size buffer.

\begin{equation}
S_i =
\begin{cases}      \text{GQC(}V_i\text{,}r \text{)} & \text{if } i = 0 \\
  \text{GQC(}S_{i-1} \oplus V_i\text{,} r \text{)} & \text{if } i > 0
\end{cases}
\label{eq:streaming_context}
\end{equation}

The SCM operates on a continuous, uninterrupted loop. While the VLM performs auto-regressive generation upon receiving a specific user query, the SCM continuously updates at each time step $T_i$ as the video stream is processed.

\subsubsection{General Query Compression}
In this section we elaborate the General Query Compression (GQC) strategy mentioned in \cref{eq:streaming_context}. GQC selects vision tokens from $V_i$ based on their text-vision attention scores. This process is query-independent, as GQC relies on a static, general purpose prompt such as "Describe the image", to calculate these attention scores, rather than using a specific query from the user. 

The full input token $I_i$ is formed by concatenating the visual tokens $V_i$ and the general query tokens $Q_g$. Following \cref{eq:attention}, the attention scores of VLM input tokens can be obtained:
\begin{equation}
\text{Attn} = \text{softmax}\left( \frac{(W_Q I_i) (W_K I_i)^T}{\sqrt{d_k}} \right) (W_V I_i)
\label{eq:attention}
\end{equation}
where $W_Q$, $W_K$, and $W_V$ are the query, key and value weight matrix and $d_k$ is the vector of dimension. 

Given a compression ratio $r$, only the tokens with the highest attention scores are kept to maintain the desired sequence length.

\subsubsection{Position Embedding Management Strategy}
VLMs often use Rotary Position Encoding (RoPE) \cite{su2023roformerenhancedtransformerrotary} to model the relative spatial and temporal relationships between visual tokens. However, visual token compression can disrupt or discard this critical positional information. To preserve these spatial and temporal dependencies post-compression, we propose to maintain the original positional embeddings for the selected visual tokens. 
In addition, processing streaming video of arbitrary length presents a positional encoding challenge, as the indices for new frames continually increase. To manage this, we track the cumulative video length at each timestep. We introduce a positional offset $P$, representing the total count of previously processed frames, and add this offset to the local position indices of the current streaming input. This method ensures that the absolute temporal position of every frame is correctly registered through the position encoding.

\subsection{Retrieved Vision Memory}

Building upon our SCM design for processing efficiency, we further propose RVM to enhance the VLM's long context retention capabilities. RVM selectively retrieves and "replays" query-relevant video clips by appending them to the model's input \cref{fig:main_method}.

The core of RVM is a memory-efficient retrieval mechanism. Each video clip at timestep $T_i$ is represented by a single key vector $K_{avg}^i$. This is achieved by average-pooling the key vectors of its visual tokens $V_i$ across the sequence length. These $K_{avg}^i$ vectors are stored in the RVM buffer, which discards the oldest entries when its maximum length is exceeded.

For retrieval, a matching query vector is generated. The query vectors $Q_{Q_s}$ from the input query tokens $Q_s$ are similarly average-pooled into a single representation, $Q_{avg}^{s}$. 

To find the most relevant historical clips, we then calculate the cosine similarity between this pooled query vector $Q_{avg}^{s}$ and each of the $n$ buffered key vectors $K_{avg}^i$ as shown in \cref{eq:similarity}.
\begin{equation}
\label{eq:similarity}
    s_i = \frac{Q_{avg}^{s} \cdot K_{avg}^i}{\|Q_{avg}^{s}\|  \|K_{avg}^i\|}
\end{equation}
where $i=0...n$, $s_i$ is the resulting similarity score for the $i$-th video clip.

Finally, the video clip corresponding to the highest similarity score is retrieved and appended to the current input stream, thereby recapturing key contextual information.

\section{Experiments}
\label{Experiments}

\begin{table*}[t]
\centering
\setlength{\tabcolsep}{4pt}
\small
\begin{tabular}{l c cccc cccc}
\toprule
\multirow{2}{*}{Model} &
\multirow{2}{*}{Frames} &
\multicolumn{4}{c}{\makecell{StreamingBench \\
Contextual Understanding}} &
\multicolumn{4}{c}{\makecell{OVOBench \\
Backward-Tracing}} \\
\cmidrule(lr){3-6}
\cmidrule(lr){7-10}
& & SQA & MCU & ACU & Avg &EPM & ASI & HLD & Avg \\
\midrule
\multicolumn{10}{c}{Streaming Video VLMs (training-based)} \\
\midrule
VideoLLM-online-8B 
& 2 fps & 30.8 & 29.2 & 24.2 &
28.0 & 22.2 & 18.8 & 12.2 &17.7 \\

FlashVStream-7B
& 1 fps & 24.8 & 25.2 & 26.8 & 25.6 &39.1 &37.2 &5.9 &27.4 \\

TimeChat-Online-7B
& 1 fps & 41.6  & 31.6 & 39.2 & 37.5 &
\underline{55.9}
&59.5 &9.7 &41.7 \\

Dispider
& 1 fps & 34.8 & 27.7 &
\underline{39.6} & 34.0 & 48.5 &55.4 &4.3 &36.1 \\

\midrule
\multicolumn{10}{c}{Streaming Video VLMs (training-free)} \\
\midrule

ReKV-LLaVA-OV-7B
& 0.5 fps & 30.4 & 28.0 & 31.2
& 29.9 &54.2 &55.7
&22.8 &44.2 \\

LiveVLM-7B
& 0.5 fps & 32.0 &
36.8 & 38.8 &35.9 & -- & -- & -- & -- \\

StreamKV-7B
& 0.5 fps & 33.2 & 30.0 & 43.6
& 35.6 &-- &--
&-- &-- \\

OASIS-Qwen2.5-VL-7B  
& 1 fps & 45.6 & 34.2 & 37.4
& 39.1 &48.8 &58.9 &30.9 & 46.2\\

\midrule

Qwen2.5-VL-7B
& 1 fps & 43.6 & 25.2 & 24.8 & 31.2 & 43.8 &57.4 &22.6 &41.3 \\

\quad + ReMem (Ours)
& 1 fps 
& \textbf{45.6} \up{2.0}
& \textbf{36.8} \up{11.6}
& \textbf{44.8} \up{20.0}
& \textbf{42.4} \up{11.2}
& \textbf{44.3} \up{0.5}
& \textbf{60.8} \up{3.4}
& \textbf{29.6} \up{7.0}
& \textbf{44.9} \up{3.6}
\\

\midrule
LLaVA-OV-7B
& 1 fps & 31.6 & 33.2 &35.6 & 33.5 & 54.2 &55.4 &21.5 & 43.7 \\

\quad + ReMem (Ours) 
& 1 fps 
& \textbf{40.4} \up{8.8}
& \textbf{37.2} \up{4.0}
& \textbf{42.0} \up{6.4}
& \textbf{36.8} \up{5.2}
& \textbf{56.0} \up{1.8}
& \textbf{57.7} \up{2.3}
& \textbf{22.9} \up{1.4}
& \textbf{45.5} \up{1.8}
\\
\midrule
Qwen3-VL-8B 
& 1 fps & \underline{46.0} & \underline{39.6} &36.4 & \underline{40.7} & 50.7 &\underline{59.7} &\underline{41.5} & \underline{50.6} \\

\quad + ReMem (Ours) 
& 1 fps 
& \textbf{47.2} \up{1.2}
& \textbf{41.0} \up{1.4}
& \textbf{42.5} \up{6.1}
& \textbf{43.6} \up{2.9}
& \textbf{52.9} \up{2.2}
& \textbf{60.8} \up{1.1}
& \textbf{43.4} \up{1.9}
& \textbf{52.4} \up{1.8}
\\
\bottomrule 
\end{tabular}
\caption{Benchmark comparison on StreamingBench and OVOBench online video evaluation dataset. }

\label{tab:online_benchmark}
\end{table*}

\begin{table*}[t]
\centering
\small
\begin{tabular}{l c cc cccc}
\toprule
\multirow{2}{*}{Model} & \multirow{2}{*}{Frames}  & \multicolumn{2}{c}{EgoSchema} & \multicolumn{4}{c}{VideoMME}\\
\cmidrule(lr){3-4} 
\cmidrule(lr){5-8}
& & Subset$^{\dagger}$ & Fullset & short & medium & long & Avg\\
\midrule

\multicolumn{8}{c}{Streaming Video VLMs (training-based)} \\
\midrule

VideoLLM-online-8B 
& 2 fps & -- & 32.8  & -- &-- &-- &--\\

FlashVStream-7B
& 1 fps & 38.40 & \textbf{68.2} & 72.0 & 61.1 & 50.3 &61.2 \\

TimeChat-Online-7B
& 1 fps & -- & 61.9 & 67.4 & 50.5 & 41.7 &53.2 \\

Dispider
& 1 fps & -- & 55.6 &66.1 &53.7 &49.7 &56.5 \\

\midrule
\multicolumn{8}{c}{Streaming Video VLMs (training-free)} \\
\midrule

ReKV-LLaVA-OV-7B
& 0.5 fps  & 64.20 & 60.70 & 69.7 &55.4 &49.3 &58.1  \\

LiveVLM-7B
& 0.5 fps & -- & 59.00 & 66.7 & 56.4 &48.8 & 57.3\\

\midrule

Qwen2.5-VL-7B
& 1 fps & 61.4 & -- & 72.1 & 66.9 & 55.4 & 64.8 \\

\quad + ReMem (Ours) 
& 1 fps 
& \textbf{62.00} \up{0.6}
& -- 
& \textbf{72.4} \up{0.1} 
& \textbf{66.9} \up{0.0} 
& \textbf{56.3} \up{0.9} 
& \textbf{65.2} \up{0.4} \\

\midrule

LLaVA-OV-7B
& 1 fps  & 61.0 & -- & 69.7 & 54.4 & 50.0 & 58.0\\

\quad + ReMem (Ours) 
& 1 fps 
& \textbf{63.4} \up{2.4}
& -- 
& \textbf{70.4} \up{0.7}
& \textbf{55.0} \up{0.6}
& \textbf{50.2} \up{0.2}
& \textbf{58.5} \up{0.5}\\
\midrule
Qwen3-VL-8B
& 1 fps  & \underline{65.4} & -- & \textbf{79.7} & \underline{70.2} & \underline{62.5} & \underline{70.8}\\

\quad + ReMem (Ours) 
& 1 fps 
& \textbf{67.0} \up{1.6}
& -- 
& \underline{78.8} \down{0.9}
& \textbf{70.9} \up{0.7}
& \textbf{62.9} \up{0.4}
& \textbf{70.9} \up{0.1}\\
\bottomrule 

\end{tabular}
\caption{Benchmark comparison on EgoSchema and VideoMME offline video evaluation dataset.$^{\dagger}$Full-set evaluation is unavailable due to evaluation server issue; results are reported on the EgoSchema subset.}

\label{tab:offline_benchmark}
\end{table*}

\subsection{Experiment Setup}
\noindent\textbf{Models and Benchmarks.} 
The ReMem framework acts as a plug-and-play adaptation for existing VLMs, enabling offline models to process streaming videos of arbitrary length. To verify the generality of our method, we apply ReMem on two widely used offline VLM baselines: LLaVA-OneVision-7B \cite{li2024llava} and Qwen2.5-VL-7B\cite{bai2025qwen2}, and conduct experiments on both online streaming tasks and offline general video understanding benchmarks. 

For online video understanding tasks, we choose commonly used StreamingBench~\cite{lin2024streamingbench} and OVOBench~\cite{li2025ovobench} evaluation benchmarks. Since ReMem is designed to improve the long context retention capability of memory based models, we specifically focus on contextual understanding tasks from StreamingBench (SQA, MCU and ACU subset) and OVOBench (EPM, ASI, HLD subset), which challenge models to precisely recall previous visual context and historical frames during streaming video intake. We report the results of online video understanding tasks in \cref{tab:online_benchmark}.

For general offline video understanding task, we choose EgoSchema~\cite{mangalam2023egoschemadiagnosticbenchmarklongform} and VideoMME~\cite{fu2025videommefirstevercomprehensiveevaluation}.
EgoSchema provides a very long-form multiple-choice video QA benchmark, containing over 5,000 human-curated question answer pairs across more than 250 hours of egocentric video, while VideoMME serves as a high-quality, full-spectrum multimodal evaluation benchmark for video analysis, 
covering clips from short snippets to hour-long videos. We follow the official evaluation protocols for each benchmark and report the results in \cref{tab:offline_benchmark}.

\noindent\textbf{Baselines and State-of-the-Arts (SOTA).}
We compare our method with a set of representative Video-LLMs covering both offline and online settings, including Qwen3-VL-8B~\cite{bai2025qwen3vltechnicalreport}, Qwen2.5-VL-7B~\cite{bai2025qwen2}, LLaVA-OneVision-7B~\cite{li2024llava} VideoLLM-Online-8B~\cite{VideoLLM-online}, FlashVStream-7B~\cite{zhang2025flashvstream}, ReKV-7B~\cite{di2025rekv}, StreamKV-7B~\cite{chen2025streamkvstreamingvideoquestionanswering}, OASIS~\cite{liang2026oasis}, Dispider~\cite{qian2025dispiderenablingvideollms}, LiveVLM~\cite{ning2025livevlmefficientonlinevideo}, and TimeChat-Online~\cite{yao2025timechat}. All models are evaluated under identical hardware platform (single H100) and decoding protocols for a fair comparison.

\subsection{Main Results}
As summarised in \cref{tab:online_benchmark} and \cref{tab:offline_benchmark}, ReMem consistently outperforms both offline baselines and streaming-oriented Video-LLMs on StreamingBench\cite{lin2024streamingbench} contextual understanding tasks and OVOBench backward tracing tasks, while remaining competitive on general video understanding EgoSchema and VideoMME. Notably, despite being training-free, ReMem achieves a 20.0-point increase in the ACU evaluation and an 11.6-point increase in MCU when compared to the Qwen2.5-VL-7B baseline. This strong performance, especially on contextual understanding tasks, validates the effectiveness of our design in improving the long-term context retention of the base model. Being a plug-and-play training-free adaptation, ReMem achieves comparable results event compared to training-based strategies. This further proves the effectiveness of our design. Furthermore, ReMem consistently improves performance on both streaming benchmarks across the Qwen2.5-VL-7B, Qwen3-VL-8B, and LLaVA-OneVision-7B base models, indicating the generalisability of our method. 

Although ReMem was designed for streaming tasks, it also generalises well to offline long-video understanding benchmarks. ReMem  outperforms its baseline on both EgoSchema and VideoMME evaluation subset. In particular, ReMem  outperforms the LLaVA-OneVision-7B base model by 2.4 points despite being designed for purely streaming task purposes. It also achieves comparable performance compared to its base models for longer video settings with VideoMME evaluation, which shows that our streaming memory design does not break the contextual inference for long contextual offline video.    

\subsection{Probing Experiments Analysis}
\label{probing experiment2}
To further validate ReMem's effectiveness in long-term context retention, we evaluated it using the visual NIAH probing experiment. We compared ReMem against the Qwen2.5-VL-7B baseline and show the result in \cref{fig:vhs_comparison}. ReMem achieves consistently higher accuracy than the baseline model across nearly all input image size settings. This finding suggests that our fixed-size memory module does not introduce the pathological long-term context loss seen in other streaming VLMs. Instead, our method successfully preserves long-term context, performing on par with or slightly better than the strong, non-streaming baseline model.

\subsection{Further Experiments}
\label{Ablation Experiments}
\subsubsection{Method Ablation}
To evaluate the effectiveness of each component, we conduct ablation experiments on the SQA and ACU tasks of StreamingBench \cite{lin2024streamingbench}. We analyse the contributions of the SCM, the RVM, and the position embedding management strategy for the SCM.

The results, reported in \cref{tab:method_ablation}, show that all components contribute positively over the Qwen2.5-VL-7B baseline. Deploying either SCM or RVM individually boosts performance on streaming tasks. Furthermore, our position embedding management strategy is also effective, enhancing performance compared to using continuous position indices post-compression. Finally, applying all proposed strategies yields the highest performance gain, indicating that our design choices are both effective and complementary.

\begin{table}[H]
\centering
\scriptsize

\begin{tabular}{c ccc cc}
\toprule
\multirow{2}{*}{Base Model} & \multicolumn{3}{c}{ReMem Components} & \multicolumn{2}{c}{StreamingBench} \\
\cmidrule(lr){2-4} \cmidrule(lr){5-6}
& SCM & Pos. Emb Mgmt. & RVM & SQA & ACU \\
\midrule

Qwen2.5-VL-7B & \xmark & \xmark & \xmark & 36.9 & 34.8 \\   
Qwen2.5-VL-7B & \cmark & \xmark & \xmark & 41.7 & 37.6 \\   
Qwen2.5-VL-7B & \xmark & \xmark & \cmark & 37.2 & 40.5 \\   
Qwen2.5-VL-7B & \cmark & \cmark & \xmark & 45.0 & 38.4 \\   
Qwen2.5-VL-7B & \cmark & \cmark & \cmark & \textbf{45.6} & \textbf{44.8} \\  

\bottomrule
\end{tabular}
\caption{Ablation of memory components on StreamingBench.}
\label{tab:method_ablation}

\end{table}
\subsubsection{Hyperparameter Study}
The performance of ReMem is influenced by key hyperparameters within its SCM and RVM modules. We first analyse the impact of these parameters by examining the chunk size, defined as the number of video frames processed in each new clip $V_i$. To isolate this effect, we evaluated the SQA and ACU subsets of StreamingBench \cite{lin2024streamingbench} using only the SCM algorithm without retrieving additional video clips via the RVM, while varying the chunk size.

As shown in \cref{tab:chunking_analysis}, model performance is highly sensitive to this setting. At a chunk size of 1, the SCM's accuracy on both SQA and ACU subtasks is at its lowest, falling even below the Qwen2.5-VL-7B baseline. Performance improves smoothly as the chunk size increases, peaking at chunk size of 16. This strong positive correlation suggests that the SCM benefits from processing longer, more context-rich clips at each update iteration. This likely allows it to better capture temporal information between adjacent frames. In contrast, processing frames individually with $\text{chunk size} = 1$ results in inferior performance, indicating that GQC compression operates more effectively when leveraging medium-sized inter-frame context.

We further evaluate the influence of the number of historical frames retrieved on the overall performance in our RVM design. We vary the number of query-relevant historical clips retrieved by ReMem using the Qwen2.5-VL-7B base model and present the results on StreamingBench in \cref{tab:retrieval_number}. We observe that the best performance is achieved when only a single clip is retrieved. While counter-intuitive, this observation aligns with recent findings in SimpleStream \cite{shen2026simplebaselinestreamingvideo}, which demonstrate that simply selecting the most recent frames outperforms many baselines. This suggests that additional historical context can introduce distracting noise or exceed the ``memory-perception'' trade-off threshold of current models.

\begin{table}[t]
\centering
\small 
\begin{tabular}{c c c c c c}
\toprule
\multirow{2}{*}{Base Model} & \multirow{2}{*}{Method}  & \multirow{2}{*}{Chunk Size}  & \multicolumn{2}{c}{StreamingBench} \\
\cmidrule(lr){4-5} 
& & & SQA & ACU \\
\midrule
Qwen2.5-VL-7B & Official & -- & 36.9 &34.8 \\
Qwen2.5-VL-7B & SCM & 1  & 35.3 &31.0\\
Qwen2.5-VL-7B & SCM & 2 & 38.6 &32.3\\
Qwen2.5-VL-7B & SCM & 4 & 41.8 &37.6 \\
Qwen2.5-VL-7B & SCM & 16 &\textbf{45.0} &\textbf{38.4}\\ 
Qwen2.5-VL-7B & SCM & 24 &42.0 &33.2\\ 

\bottomrule
\end{tabular}
\caption{Video input chunk size analysis for Streaming Context Memory (SCM)}
\label{tab:chunking_analysis}
\end{table}

\begin{table}[h]
\centering
\renewcommand{\arraystretch}{1.2} 
\setlength{\tabcolsep}{12pt} 

\begin{tabular}{lccc}
\toprule
\textbf{Number of Clips} & \textbf{SQA} & \textbf{MCU} & \textbf{ACU} \\
\midrule
1 & \textbf{45.6} & \textbf{36.8} & \textbf{44.8} \\
2 & 41.6 & 34.0 & 41.2 \\
3 & 41.6 & 29.2 & 38.2 \\
\bottomrule
\end{tabular}
\caption{Influence of the number of retrieved historical clips on overall performance within the RVM design.}
\label{tab:retrieval_number}
\end{table}

\subsubsection{Efficiency Analysis}
In this section we present a latency and memory consumption study of ReMem compared against some other baseline and SOTA streaming VLM models. 

\begin{figure}[t]
  \centering
\includegraphics[width=1\linewidth]{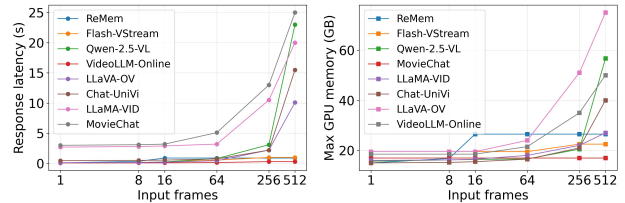}
   
   \caption{Left: response latency of different VLM models under streaming evaluation. Right: Maximum GPU memory consumption of different VLM models under streaming evaluation.}
   \label{fig:latency_memory}
\end{figure}

 We conduct both latency and memory experiments on StreamingBench \cite{lin2024streamingbench} SQA subset for better evaluation efficiency. We select parameter setting of compression ratio $r = 0.7$, $ \text{chunk size} = 16$ and $ \text{fps} = 1$. Our experiments are conducted on a single Nvidia 80GB H100 GPU. As illustrated in \cref{fig:latency_memory}, ReMem maintains a constant latency of 1.0 second even as the number of input video frames scales to 512. This stability is a direct result of our chunk-based streaming architecture and fixed memory allocation. Furthermore, ReMem’s memory footprint plateaus at 32 GB once the buffer is fully utilized, demonstrating high hardware efficiency for processing arbitrarily long video streams.
\section{Conclusion}
In this paper, we addressed a key limitation in existing streaming VLMs: the progressive loss of long-term context in fixed-size memory architectures. To tackle this, we proposed ReMem, a novel, training-free adaptation strategy designed to achieve streaming video understanding with long context retention. ReMem's dual-memory architecture features a Streaming Context Memory (SCM) for continuous, query-independent visual token compression and a Retrieved Vision Memory (RVM) for query-specific salient vision information augmentation. 

Our comprehensive experiments on online evaluation benchmarks and general long-video benchmarks demonstrated that ReMem significantly outperforms both baseline models and competing memory-based methods. Crucially, our (NIAH) probing experiments confirmed that ReMem successfully preserves long-term context, avoiding the pathological information loss observed in other streaming approaches. As a flexible, training-free method, ReMem enables the deployment of pretrained VLMs in, real-world applications. 
{
\small
\bibliography{aaai2027}
}
\clearpage
\onecolumn
\appendix

\begin{center}
{\Large\textbf{Supplementary Material for ReMem: Streaming Video Understanding With Long Context Retention}}
\end{center}

\section{Additional Experimental Details}

In this section, we present additional experimental results from our streaming benchmark evaluations and conduct further analyses to verify the effectiveness of our design choices.

\subsection{Full Streaming Benchmark Results}

The proposed training-free ReMem framework is designed to preserve context and mitigate information loss in long streaming video inputs. While our primary evaluation focused on contextual reasoning, specifically the Contextual Understanding subtask in StreamingBench and the Backward-Tracing subtask in OVOBench, we also report comprehensive results across both benchmarks to demonstrate our method's versatility in other settings. The Real-Time Visual Understanding results for StreamingBench are detailed in \cref{tab:streamingbench_realtime}, and the Real-Time Visual Perception results for OVOBench are provided in \cref{tab:OVOBench realtime}.

\begin{table*}[h]
\centering
\small
\renewcommand{\arraystretch}{1.1}
\setlength{\tabcolsep}{5pt} 
\begin{tabular}{lr cccccccccc c}
\toprule
\textbf{Model} & \textbf{\#Frames} & \textbf{OP} & \textbf{CR} & \textbf{CS} & \textbf{ATP} & \textbf{EU} & \textbf{TR} & \textbf{PR} & \textbf{SU} & \textbf{ACP} & \textbf{CT} & \textbf{Avg.} \\
\midrule
\multicolumn{13}{c}{\textbf{Streaming Video VLMs (training-based)}} \\
\midrule
VideoLLM-online-8B   & 2 fps   &39.1 &40.1 &34.5 & 31.1 &46.0 & 32.4 &31.5 &34.2 & 42.5 & 27.9 & 36.0\\
FlashVStream-7B      & 1 fps   &25.9 &43.6 &24.9 &23.9 &27.3 & 13.1& 18.5& 25.2& 23.9& 48.7&23.2 \\
TimeChat-Online-7B   & 1 fps   & 80.2&82.0 &79.5 &83.3 &76.1 &78.5 &78.7 &64.6 &\textbf{69.6} &\textbf{58.0} &75.4 \\
Dispider             & 1 fps   & 74.9& 75.5& 74.1&73.1 &74.4 &59.9 &76.1 &62.9 &62.2 &45.8 &67.6 \\
\midrule
\multicolumn{13}{c}{\textbf{Streaming Video VLMs (training-free)}} \\
\midrule
ReKV-LLaVA-OV-7B     & 0.5 fps &76.0 &\textbf{81.3} &77.9 &76.9 &66.0 &66.0 &69.4 &61.0 &64.3 &49.2 &69.2 \\
LiveVLM-7B           & 0.5 fps &81.5 &78.1 &83.3 &79.1& 69.6&74.1 &75.0 &\underline{69.1} &67.7 &40.4 &72.9 \\
StreamKV-7B          & 0.5 fps &73.8 &77.3 &85.9 &77.5 &73.3 &63.9 &69.4 &61.4 &63.2 &35.8 &68.8 \\
\midrule
Qwen2.5-VL-7B        & 1 fps & 77.0 & \underline{79.7} & 78.9 & 81.1 & 74.8 & 76.6 & 74.1 & 64.6 & 66.2 & 54.8 & 73.2 \\
\quad + ReMem (Ours)$^{\dagger}$ & 1 fps & 75.9 & 70.3 & 78.5 & 80.1 & 73.0 & 76.3 & 76.9 & 64.2 & 64.5 & 44.7 & 71.3 \\
\midrule
LLaVA-OV-7B          & 1 fps & 81.8 & 78.9 & 82.3 & 82.1 & 71.7 & 72.0 & 71.3 & 66.3 & 67.6 & 44.7 & 73.1 \\
\quad + ReMem (Ours)$^{\dagger}$ & 1 fps & 82.7 & 72.7 & 81.4 & \underline{84.3} & 73.0 & 71.0 & 71.3 & 64.6 & 72.4 & 38.8 & 73.1 \\
\midrule
Qwen3-VL-8B          & 1 fps   &\underline{85.6} &78.2 &\underline{86.1} &\textbf{86.3} &\textbf{79.0} &\textbf{90.4} &\underline{81.7} &69.0 &67.3 &\underline{56.5} &\underline{78.0} \\
\quad + ReMem (Ours)$^{\dagger}$ & 1 fps   &\textbf{85.9} &79.3 &\textbf{86.4} &\textbf{86.3} &\underline{78.2} &\underline{90.1} &\textbf{81.9} &\textbf{69.4} &\underline{68.0} &56.1 &\textbf{78.2} \\
\bottomrule 
\end{tabular}
\caption{Detailed performance comparison of VLMs across StreamingBench real-time evaluation.  All results are accuracies in
percent. The Overall average is computed over
all ten tasks. For rows marked with $\dagger$, each task reports the
best result across the evaluated ReMem chunk sizes: $\{4,8,16,32\}$ for
Qwen2.5-VL-7B, $\{4,8,16\}$ for LLaVA-OV-7B and $\{4,8\}$ for Qwen3-VL-8B.}
\label{tab:streamingbench_realtime}
\end{table*}

\begin{table*}[h]
\centering
\small
\renewcommand{\arraystretch}{1.1}
\setlength{\tabcolsep}{4.5pt}
\begin{tabular}{lc cccccc c ccc c c}
\toprule
\multirow{2}{*}{\textbf{Model}}
& \multirow{2}{*}{\textbf{\#Frames}}
& \multicolumn{7}{c}{\textbf{Real-Time Visual Perception}}
& \multicolumn{4}{c}{\textbf{Backward Tracing}}
& \textbf{Overall} \\
\cmidrule(lr){3-9}
\cmidrule(lr){10-13}
\cmidrule(lr){14-14}
& & \textbf{OCR} & \textbf{ACR} & \textbf{ATR}
& \textbf{STU} & \textbf{FPD} & \textbf{OJR} & \textbf{Avg.}
& \textbf{EPM} & \textbf{ASI} & \textbf{HLD} & \textbf{Avg.}
& \textbf{Avg.} \\
\midrule

\multicolumn{14}{c}{\textbf{Streaming Video VLMs (training-based)}} \\
\midrule
VideoLLM-online-8B
& 2 fps
& 8.1 & 23.9 & 12.1 & 14.0 & 45.5 & 21.2 & 20.8
& 22.2 & 18.8 & 12.2 & 17.7 & 19.3 \\

FlashVStream-7B
& 1 fps
& 25.5 & 32.1 & 29.3 & 33.7 & 29.7 & 28.8 & 29.9 & 36.4 & 33.8 & 5.9 & 25.4 & 27.6 \\

TimeChat-Online-7B
& 1 fps
& 75.2 & 46.8 & 70.7 & 47.8 & 69.3 & 61.4 & 61.9 & 55.9 & 59.5 & 9.7 & 41.7 & 51.8 \\

Dispider
& 1 fps
& 57.7 & 49.5 & 62.1 & 44.9 & 61.4 & 51.6 & 54.6 & 48.5 & 55.4 & 4.3& 36.1 & 45.3 \\

LLaVA-Video-7B & 1 fps &  69.1& 58.7& 68.8& 49.4& \underline{74.3}& 59.8& 62.3& \textbf{56.2}& 53.5& \textbf{51.5}& \textbf{53.7} & 60.1 \\
\midrule

\multicolumn{14}{c}{\textbf{Streaming Video VLMs (training-free)}} \\
\midrule
ReKV-LLaVA-OV-7B
& 0.5 fps
& 52.3 &54.1 & 69.8 & 43.3 & 67.3 & 57.1 & 57.3 & 57.6 & 56.1 & 18.8 & 44.2 & 50.8 \\

OASIS-Qwen2.5VL-7B
& 1 fps
& \textbf{84.7} & \textbf{72.0} & 65.8 & 51.9 & 67.5 & 64.2 & \underline{67.7} & 48.8 &
58.9 &
30.9 &
46.2 & 60.5 \\
\midrule

Qwen2.5-VL-7B
& 1 fps
& 71.8 & 54.1 & 62.1 & 49.4 & 62.4 & 58.2 & 59.7
& 43.8 & 57.4 & 22.6 & 41.3 & 53.5 \\

\quad + ReMem (Ours)$^{\dagger}$
& 1 fps
& 67.1 & 56.0 & 63.8 & 48.9 & 60.4 & 51.1 & 57.9
& 44.3 & \textbf{60.8} & 29.6 & 44.9 & 53.4 \\
\midrule

LLaVA-OV-7B
& 1 fps
& 72.5 & 59.6 & 73.3 & 51.7 & 71.3 & \underline{64.7} & 65.5
& 54.2 & 55.4 & 21.5 & 43.7 & 58.2 \\

\quad + ReMem (Ours)$^{\dagger}$
& 1 fps
& 78.5 & \underline{70.6} & \textbf{76.7} & 55.1 & \textbf{75.2} & \textbf{76.6} & 72.1
& \underline{56.0} & 57.7 & 22.9 & 45.5 & \textbf{63.3} \\
\midrule

Qwen3-VL-8B 
& 1 fps
& 80.2 & 58.5 & \underline{74.8} & \underline{56.6} & 71.9 & 60.3 & 67.1
& 50.7 & \underline{59.7} & 41.5 & 50.6 & 61.6 \\

\quad + ReMem (Ours)$^{\dagger}$
& 1 fps
& \underline{81.3} & 59.4 & 74.1 & \textbf{57.5} & 72.7 & 59.9 & \textbf{67.5}
& 52.9 & \textbf{60.8} & \underline{43.4} & \underline{52.4} & \underline{62.4} \\
\bottomrule
\end{tabular}

\caption{
Detailed performance comparison of VLMs across OVOBench Real-Time Visual
Perception and Backward Tracing tasks. All results are accuracies in
percent. The Real-Time and Backward averages are computed over six and
three tasks, respectively, while the Overall average is computed over
all nine tasks. For rows marked with $\dagger$, each task reports the
best result across the evaluated ReMem chunk sizes: $\{4,8,16,32\}$ for
Qwen2.5-VL-7B, $\{4,8,16\}$ for LLaVA-OV-7B and $\{4,8\}$ for Qwen3-VL-8B.
}
\label{tab:OVOBench realtime}
\end{table*}

\subsection{Prompting Strategies Analysis}

\noindent\textbf{Two-stage protocol.}
To measure how closely a model preserves its task-specific attention behavior under different prompting conditions, we introduce a unified two-stage protocol using \emph{identical} visual inputs. 
Stage~1 applies StreamingBench SQA evaluation utilising specific task prompt  to obtain the reference attention distribution. 
Stage~2 then computes a second attention distribution using an alternative prompt. 
The three variants differ only in the Stage~2 prompt; all other factors (video content, decoding budget, and inference process) are kept fixed. 
Average cosine similarities between Stage~1 and Stage~2 are reported in Table~\ref{tab:avg_attn_similarity}. 

\noindent\textbf{Hooking prompt} \cite{zhang2025oasisimageneedmultimodal} examines how the model behaves when no textual guidance is provided.  
\textbf{Design.} Following the Oasis ``hooking'' setup, the second pass removes both the natural language query and the assistant side template, and feeds only the visual tokens together with the fixed pre-query template. Concretely, for Qwen2.5-VL we drop the instruction and the post query template \texttt{<|im\_end|><|im\_start|>assistant} , so the model must autoregressively generate text conditioned solely on the video, without any explicit instruction.   
\textbf{Result.} This setting yields the \emph{lowest} similarity (\textbf{82.1} avg.), indicating that removing all textual structure causes the largest deviation from task-driven attention behavior.

\noindent\textbf{Chat-template prompt}\cite{yang2025streammemqueryagnostickvcache} assesses whether providing only conversational structure without any explicit instruction helps stabilize model behavior.  
\textbf{Design.} The second pass still omits the instruction text, but preserves the full chat template boundary between user and assistant. In practice, we append the assistant header tokens \texttt{<|im\_end|><|im\_start|>assistant} directly after the visual tokens, leveraging the model’s pretraining on caption style data to implicitly elicit a generic video description even in the absence of a natural-language query.  
\textbf{Result.} This configuration produces \emph{intermediate} similarity (\textbf{85.7} avg.), showing that structural cues partially constrain the model compared to fully instruction-free prompting.

\noindent\textbf{General prompt (ours)} induces explicit, query-agnostic reconstruction behavior that closely matches the model’s specific task attention patterns.   
\textbf{Design.} The second pass pairs the video with a general yet concise instruction asking the model to “describe the video in detail sequentially from start to finish”; the standard chat template is preserved.  
\textbf{Result.} This strategy achieves the \emph{highest} alignment (\textbf{92.7} avg.), indicating that explicit textual guidance best preserves the model’s task-specific attention dynamics. 

For consistency, all the prompting experiments are conducted on the SQA subset of StreamingBench with a total of 250 question samples. Similarly we set compression ratio $r = 0.7$, $ \text{chunk size} = 16$ and $ \text{fps} = 1$ for all the experiments. We also set the "max token length" parameter during decoding to be 1 to reduce the time for attention score generation. 

The exact query-agnostic general prompt in our strategy is "You are an advanced video question-answering AI assistant. You have been provided with a video. Your task is to describe the video in detail sequentially from start to finish."

\begin{table*}[h]
\centering

\begin{minipage}[h]{0.52\textwidth}
\centering
\begin{tabular}{c c c c}
\toprule
Base Model & Method & Ratio & SQA \\
\midrule
Qwen2.5-VL-7B & Official & 1.0 & \textbf{43.6} \\
\midrule
Qwen2.5-VL-7B & VisionZip & 0.1 & 35.9 \\
Qwen2.5-VL-7B & GQC & 0.1 & 36.6\up{0.7} \\
\midrule
Qwen2.5-VL-7B & VisionZip & 0.3 & 40.4 \\
Qwen2.5-VL-7B & GQC & 0.3 & 41.0\up{0.6} \\
\midrule
Qwen2.5-VL-7B & VisionZip & 0.7 & 43.2 \\
Qwen2.5-VL-7B & GQC & 0.7 & \textbf{43.5}\up{0.3} \\
\bottomrule
\end{tabular}
\caption{Comparison of visual token compression methods on StreamingBench under different retention ratios.}
\label{tab:compression study}
\end{minipage}%
\hfill
\begin{minipage}[h]{0.45\textwidth}
\centering
\begin{tabular}{c c c}
\toprule
Prompt Strategy & Best Sim. & Avg. Sim. \\
\midrule
Hooking        & 89.5    & 82.1 \\
Chat-Template  & 92.6    & 85.4 \\
General (ours) & \textbf{98.1} & \textbf{92.7} \\
\bottomrule
\end{tabular}
\caption{Average cosine similarity between attention maps from Stage~1 Multiple Choice Questions and Stage~2 under different second-pass prompting strategies.}
\label{tab:avg_attn_similarity}
\end{minipage}

\end{table*}

\subsection{Compression Experiment}
We further analyse the hyperparameter for the compression ratio, $r$, on the StreamingBench SQA subtask. Using the Qwen2.5-VL-7B baseline, we compared our method GQC against VisionZip \cite{yang2025visionzip}. The results in \cref{tab:compression study} show a clear trade-off: performance is nearly lossless at $r=0.7$ for both methods but degrades as compression increases. With performance lowest for both at $r=0.1$, we empirically selected $r=0.7$ as the optimal balance between accuracy and compression and used it for all subsequent experiments.

\begin{figure*}[h]
    \centering
    \includegraphics[width=\textwidth]{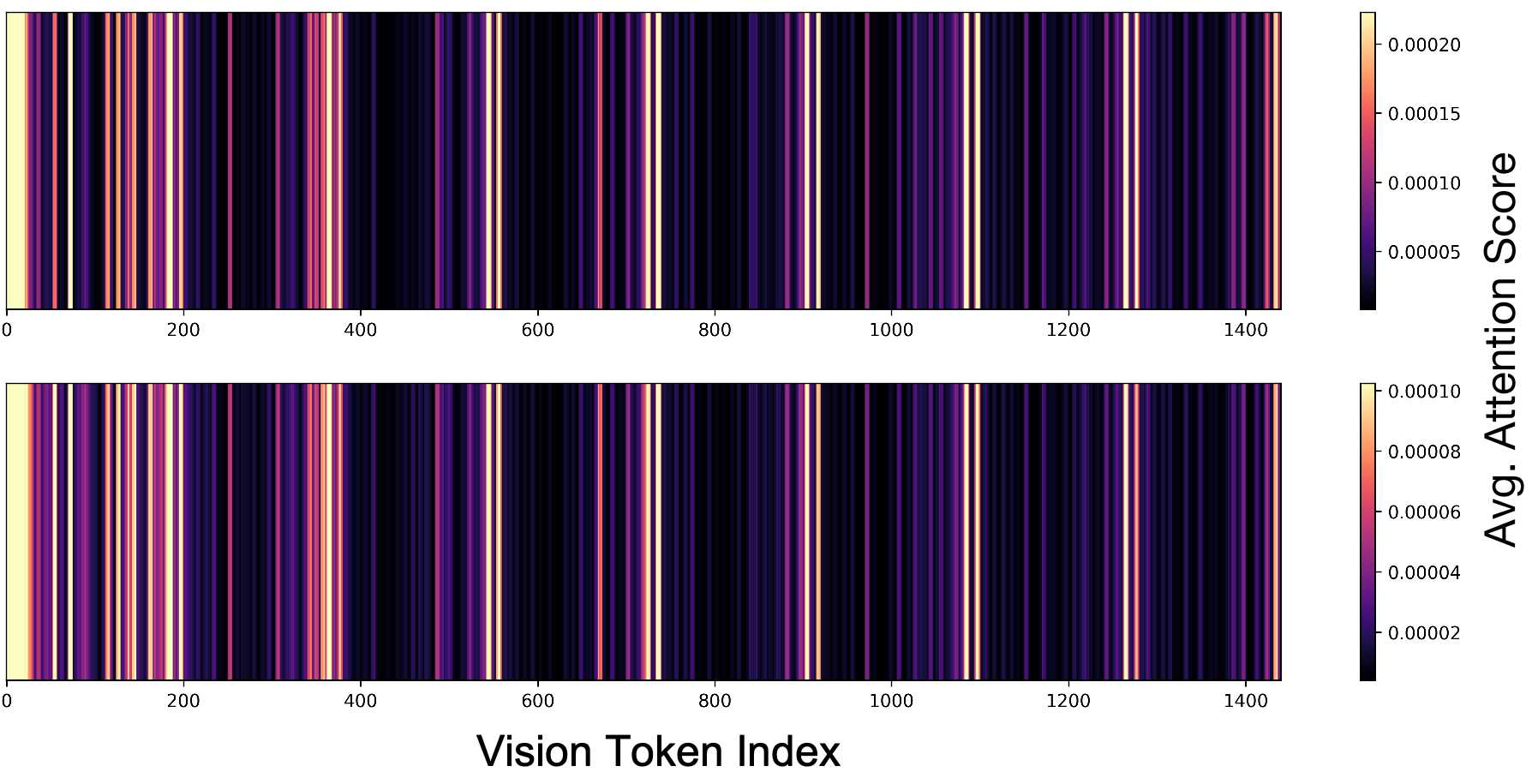}
    \caption{Distribution of attention score over visual tokens.  \textbf{Top:} general query. \textbf{Bottom:} task query.  
    The task query is \textit{``How many game scores has the team referred to in the previous question scored so far?''} from \textsc{StreamingBench} SQA, 
    and the general query is \textit{``Describe the video in detail sequentially from start to finish.''}}
    \label{fig:similarity_compare}
\end{figure*}

\cref{fig:similarity_compare} shows an illustrative example of the attention score distribution across visual tokens under different queries. While the absolute attention scores differ, the spatial distribution over visual tokens remains highly consistent across the two passes, indicating that our general query approach preserves the attention pattern of visual tokens and the general query based compression GQC works well with different task queries.  

\section{Method Elaboration}
In \cref{alg:visual compression}, we provide detailed pseudocode for selecting vision tokens based on attention scores, as introduced in our General Query Compression (GQC) module.

\begin{algorithm}[h]
\caption{Pseudo-code for attention based vision token compression}
\label{alg:topk_selection}

\begin{minipage}[t]{0.48\textwidth}
\begin{algorithmic}[1]
\REQUIRE $V_i$: input vision tokens 
\REQUIRE $Attn$: attention weights from all heads and layers 
\REQUIRE $I_{start}$: image token start index; 
\REQUIRE $I_{end}$: image token end index; 
\REQUIRE $r$: compression ratio 

\STATE $\text{Attn}_{\text{layer\_avg}} \gets \text{Mean}(Attn, \text{dim}=0)$
\item[] \COMMENT{Average across layers}
\STATE $\text{Attn}_{\text{head\_avg}} \gets \text{Mean}(\text{Attn}_{\text{layer\_avg}}, \text{dim}=0)$
\item[] \COMMENT{Average across heads}
\STATE $\mathcal{A}_{\text{text} \to \text{img}} \gets \text{Attn}_{\text{head\_avg}}[I_{end}+1 : , \quad I_{start} : I_{end}+1]$
\item[] \COMMENT{img-text attention score}
\end{algorithmic}
\label{alg:visual compression}
\end{minipage}%
\hfill
\begin{minipage}[t]{0.48\textwidth}
\begin{algorithmic}[1]
\makeatletter
\setcounter{ALC@line}{3} 
\makeatother
\STATE $\mathcal{S}_{\text{img}} \gets \text{Mean}(\mathcal{A}_{\text{text} \to \text{img}}, \text{dim}=0)$
\item[] \COMMENT{Average scores from all text queries}

\STATE $N_{\text{img}} \gets \text{Shape}(\mathcal{S}_{\text{img}})[-1]$
\item[] \COMMENT{Get total number of vision tokens}
\STATE $k \gets \lfloor r \times N_{\text{img}} \rfloor$
\item[] \COMMENT{Calculate number of tokens $k$ to select}

\STATE $Idx_{topk} \gets \text{TopK}(\mathcal{S}_{\text{img}}, k, \text{dim}=-1)$
\item[] \COMMENT{Find indices of $k$ highest scores}
\STATE $V_{i}^{'} \gets V_{i}[Idx_{topk}]$
\item[] \COMMENT{Select vision tokens}
\end{algorithmic}
\end{minipage}

\end{algorithm}

\section{Case Studies}
\begin{figure*}[h]
    \centering
    \includegraphics[width=0.88\linewidth]{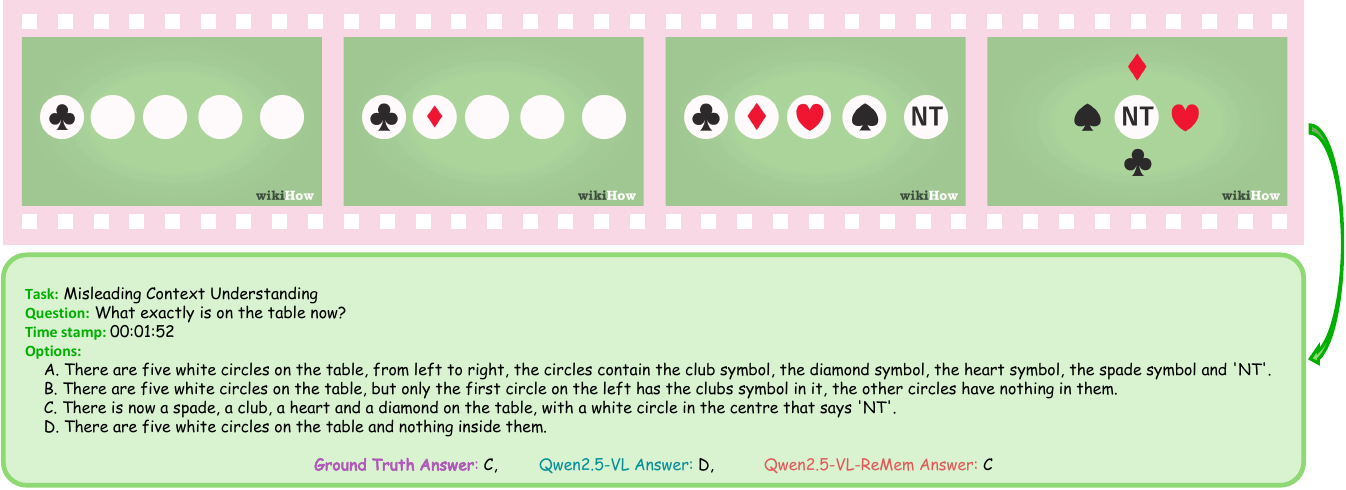}
    \caption{
    \textbf{Case study} on \textit{Misleading Context Understanding}.
    The video provides a detailed introduction to the rules and scoring of the card game Bridge, including the suit ranking: Clubs, Diamonds, Hearts, Spades, and No Trump (NT).
    At timestamp 00:01:52 the question asks: \textit{“What exactly is on the table now?”} with four multiple choice options describing different symbol configurations. 
    The ground truth answer is \textbf{C},
    while the Qwen2.5-VL baseline is distracted by earlier frames and incorrectly predicts \textbf{D}. In contrast, our ReMem-enhanced model correctly outputs \textbf{C}, highlighting its improved ability to track evolving object states over time and resist misleading intermediate context.
    }
    \label{fig:case1}
\end{figure*}

\begin{figure*}[h]
    \centering
    \includegraphics[width=0.88\linewidth]{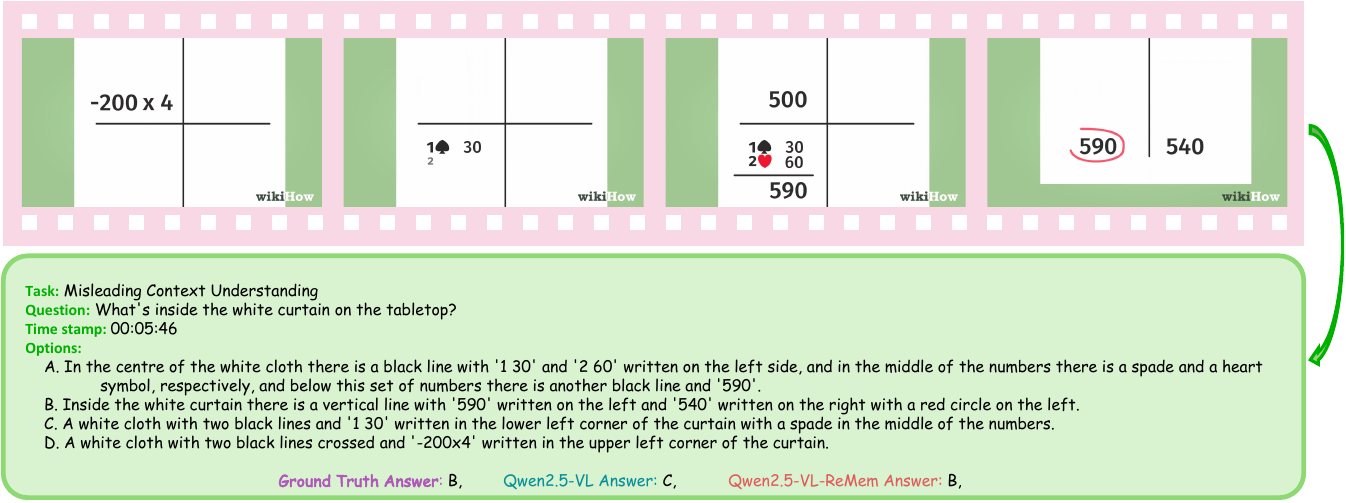}
    \caption{
    \textbf{Case study} on \textit{Misleading Context Understanding}.
    The video explains the rules and scoring of the card game Bridge, and this example shows several snapshots of the scorecard as the match progresses, ending with the final tally sheet on the tabletop.
    At timestamp 00:05:46, the question asks: \textit{“What's inside the white curtain on the tabletop?”} with four multiple choice options describing different numerical layouts. 
    The ground truth answer is \textbf{B},
    while the Qwen2.5-VL baseline is distracted by earlier frames and incorrectly predicts \textbf{C}. In contrast, our ReMem-enhanced model correctly outputs \textbf{B}, highlighting its improved ability to track evolving object states over time and resist misleading intermediate context.
    }
    \label{fig:case2}
\end{figure*}

\begin{figure*}[h]
    \centering
    \includegraphics[width=0.88\linewidth]{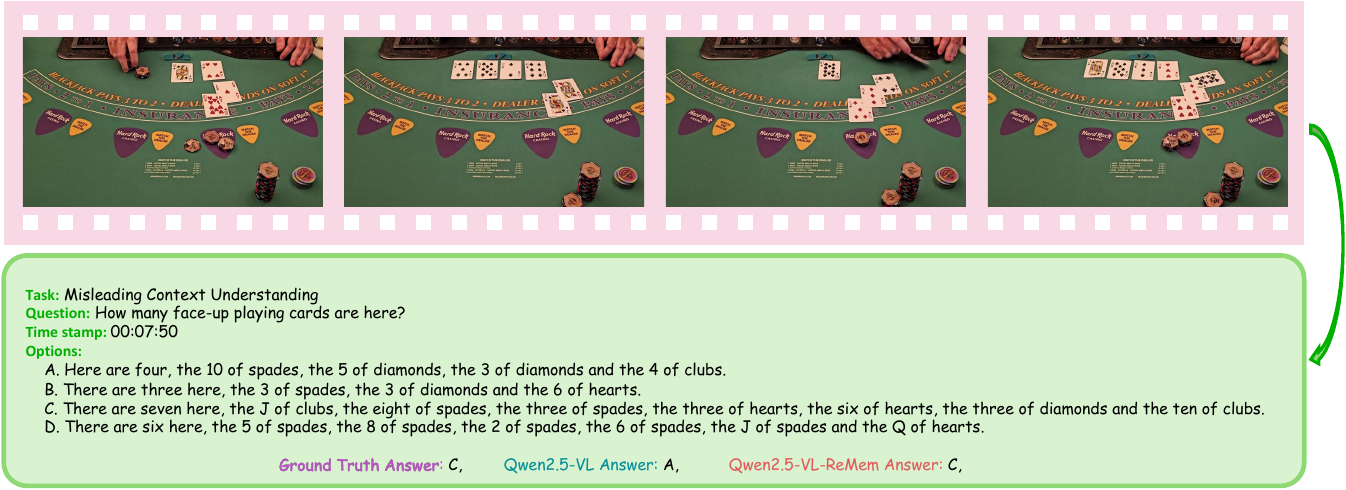}
    \caption{
    \textbf{Case study} on \textit{Misleading Context Understanding}.
    This example shows a sequence of frames from a casino table game where face-up cards are gradually revealed and rearranged.
    At timestamp 00:07:50, the question asks: \textit{“How many face-up playing cards are here?”} with four multiple choice options describing different cards arrangement.
    The ground truth answer is \textbf{C},
    while the Qwen2.5-VL baseline is distracted by earlier frames and incorrectly predicts \textbf{A}. In contrast, our ReMem-enhanced model correctly outputs \textbf{C}, highlighting its improved ability to track evolving object states over time and resist misleading intermediate context.
    }
    \label{fig:case3}
\end{figure*}

\begin{figure*}[h]
    \centering
    \includegraphics[width=0.88\linewidth]{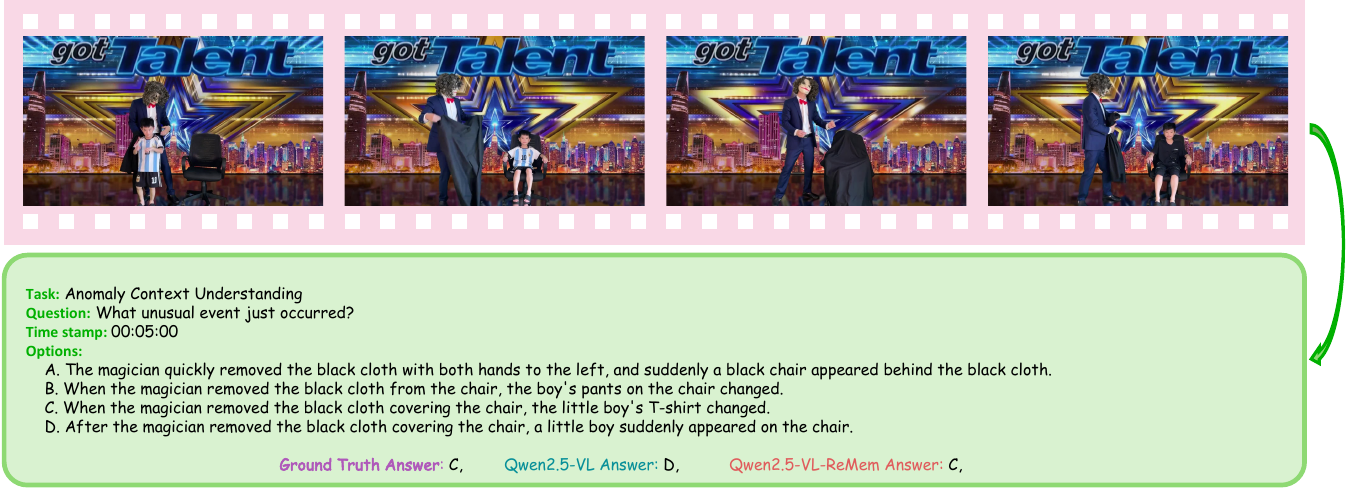}
    \caption{
    \textbf{Case study} on \textit{Anomaly Context Understanding}. 
    The video shows a stage magic performance where a magician covers a chair with a black cloth while a young boy stands beside it. 
    At timestamp 00:05:00, the question asks \textit{``What unusual event just occurred?''}, with four multiple choice options describing different outcomes after the cloth is removed. 
    The ground truth answer is \textbf{C},
    while the Qwen2.5-VL baseline incorrectly predicts \textbf{D}, hallucinating that a boy suddenly appears on the chair. 
    In contrast, our ReMem-enhanced model correctly selects \textbf{C}, indicating improved sensitivity to fine-grained, unexpected events and more precise reasoning in dynamic, unpredictable scenes. 
    }
    \label{fig:case4}
\end{figure*}

\begin{figure*}[h]
    \centering
    \includegraphics[width=0.88\linewidth]{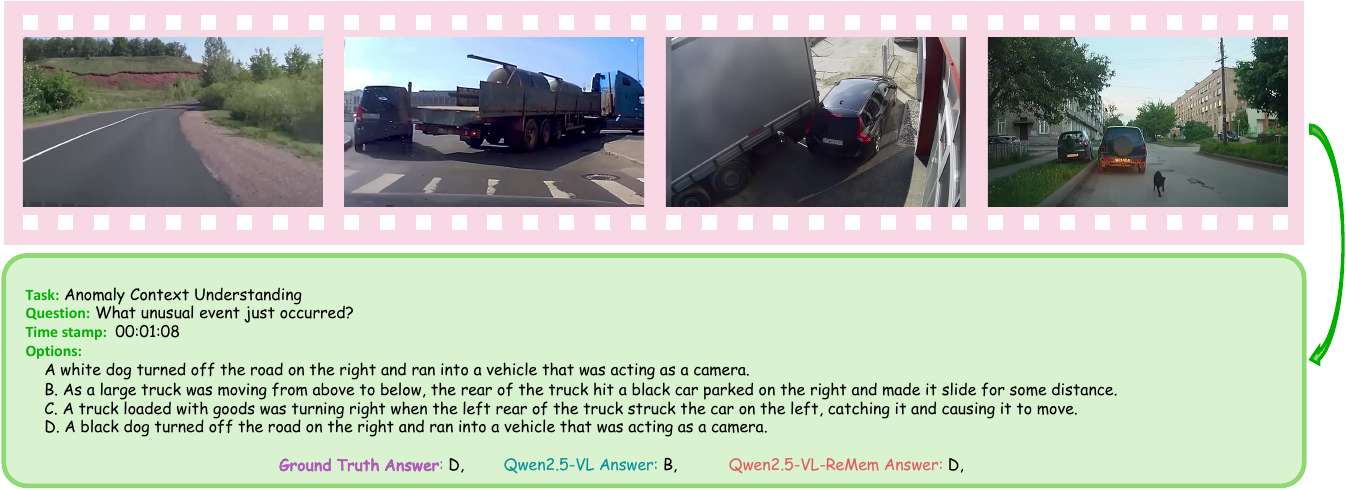}
    \caption{
    \textbf{Case study} on \textit{Anomaly Context Understanding}.  
    This video presents a dashcam-style traffic video in which rare or unexpected events must be localized and identified.
    At timestamp 00:01:08, the question asks: \textit{``What unusual event just occurred?''}, with four multi choice descriptions of possible anomalies.  
    The ground truth answer is \textbf{D},
    while the Qwen2.5-VL baseline incorrectly predicts \textbf{B}, being distracted by large moving vehicles in earlier frames. In contrast, our ReMem-enhanced model correctly identifies \textbf{D}, demonstrating stronger temporal grounding and improved resilience to misleading context in fast-changing real-world scenes.  
    }
    \label{fig:case5}
\end{figure*}

\begin{figure*}[h]
    \centering
    \includegraphics[width=0.88\linewidth]{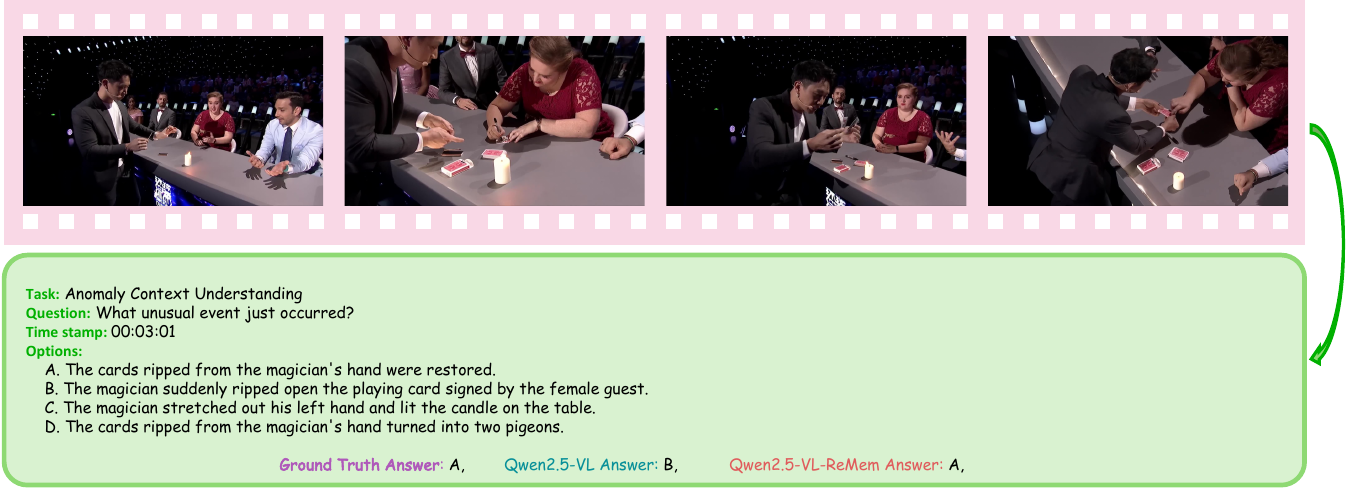}
    \caption{
    \textbf{Case study} on \textit{Anomaly Context Understanding}. 
    The video records a stage magic performance in which a magician interacts with guests at a table.
    At timestamp 00:03:01, the question asks: \textit{``What unusual event just occurred?''} with four multiple choice options describing possible outcomes. 
    The ground-truth answer is \textbf{A}, 
    while the Qwen2.5-VL baseline incorrectly predicts \textbf{B}, focusing on ripping the signed card, a salient but non-anomalous action. In contrast, our ReMem-enhanced model correctly selects \textbf{A}. 
    This example demonstrates that ReMem better captures temporal dependencies and distinguishes genuinely unexpected events from normal but visually prominent actions, leading to more reliable anomaly reasoning in streaming video.
    }
    \label{fig:case6}
\end{figure*}

\begin{figure*}[h]
    \centering
    \includegraphics[width=0.88\linewidth]{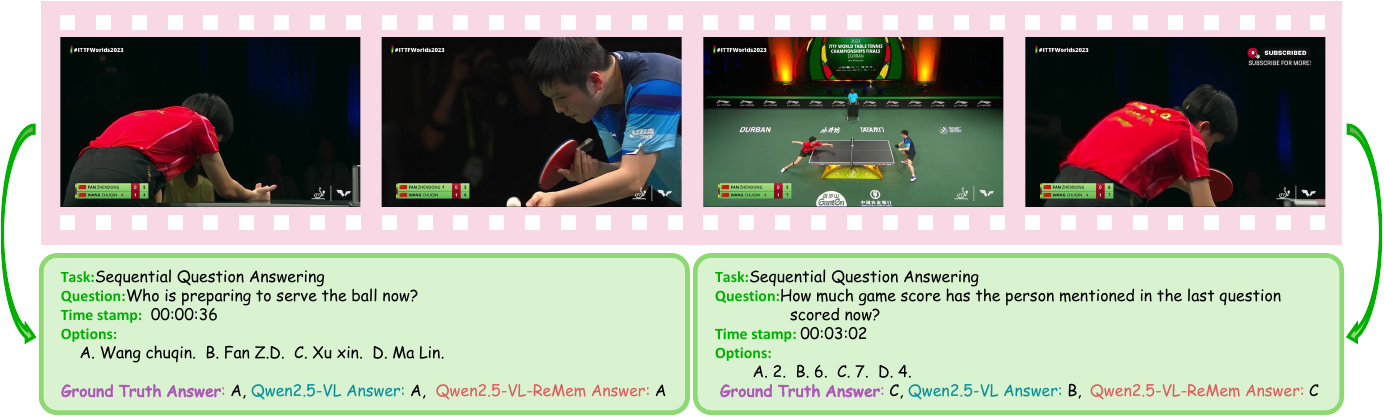}
    \caption{
    \textbf{Case study} on \textit{Sequential Question Answering}. 
    The video records a professional table-tennis match, where the model must answer five temporally linked questions based on the same ongoing rally. 
    \textbf{The first question} at timestamp 00:00:36 asks: \textit{``Who is preparing to serve the ball now?''}, and all models correctly identify \textit{Wang Chuqin} (option~A). 
    \textbf{The second question} at timestamp 00:03:02 then queries:\textit{``How much game score has the person mentioned in the last question scored now?''} with four multiple choice options describing different scores. 
    The ground truth answer is \textbf{C},while the Qwen2.5-VL baseline fails to propagate the correct entity and its evolving score,incorrectly predicting \textbf{B}. In contrast, our ReMem-enhanced model maintains consistent temporal grounding and accurately outputs \textbf{C}. This example illustrates that ReMem substantially improves long-range entity tracking and state updating in streaming video question answering.
    }
    \label{fig:case7}
\end{figure*}

\begin{figure*}[h]
    \centering
    \includegraphics[width=0.88\linewidth]{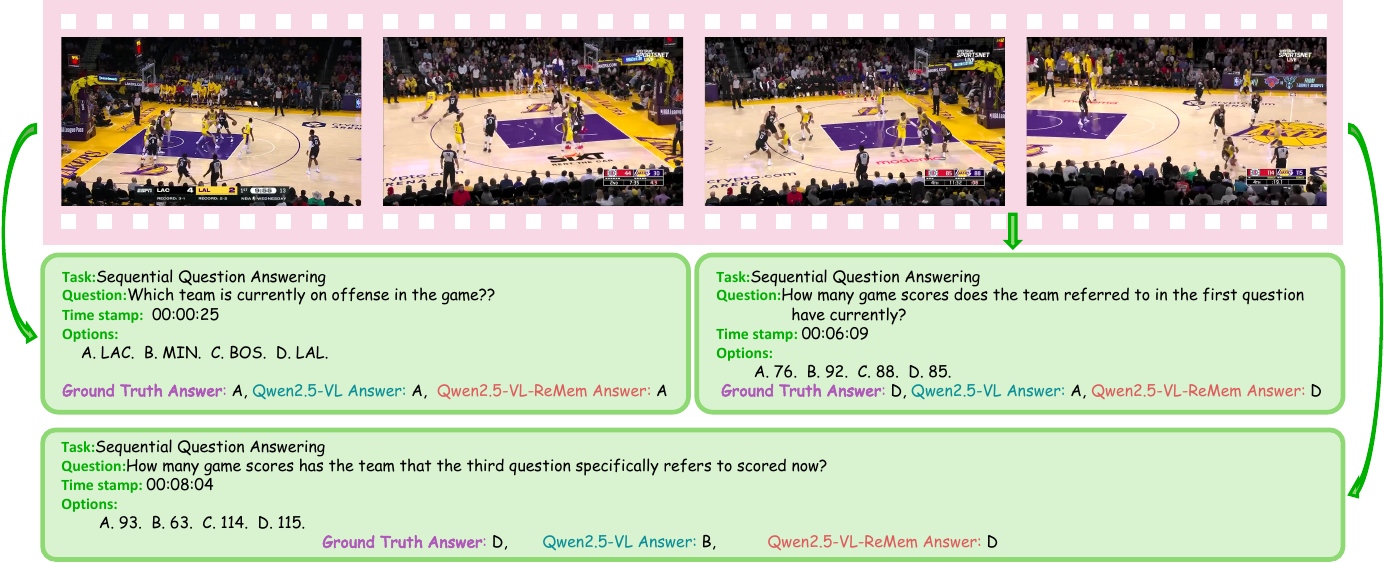}
    \caption{
    \textbf{Case study} on \textit{Sequential Question Answering (SQA)}. 
    The video shows a professional basketball game, where the model is queried with five temporally dependent questions about the same ongoing play.
    \textbf{The first question} at timestamp 00:00:25 asks: \textit{``Which team is currently on offense in the game?''}, and all models correctly identify \textit{LAC} (option~A). 
    \textbf{The fourth question} at timestamp 00:06:09, asks: \textit{``How many game scores does the team referred to in the first question have currently?''}.
    \textbf{The fifth question} at timestamp 00:08:04 further asks: \textit{``How many game scores has the team that the third question specifically refers to scored now?''}, both under a multi-choice setting.
    The ground-truth answers are \textbf{D} (85 points) and \textbf{D} (115 points), respectively. 
    The Qwen2.5-VL baseline fails to correctly update and track the scores, predicting \textbf{A} and \textbf{B}, whereas our ReMem-enhanced model consistently outputs the correct options \textbf{D} and \textbf{D}. 
    This example highlights that ReMem more effectively leverages episodic memory to maintain entity references and their evolving states over time, enabling coherent and contextually accurate reasoning across a sequence of related questions in streaming video.
    }
    \label{fig:case8}
\end{figure*}

In this section we visualise some of the benchmark questions and answers for our method and baseline method on StreamingBench \cite{lin2024streamingbench} sub tasks. The case study samples are shown in \cref{fig:case1} to \cref{fig:case8}.

\end{document}